\documentclass[10pt]{article} 

\usepackage[preprint]{rlj} 

\usepackage{amssymb}            
\usepackage{mathtools}          
\usepackage{mathrsfs}           
\usepackage{graphicx}           
\usepackage{subcaption}         
\usepackage[space]{grffile}     
\usepackage{url}                
\usepackage{lipsum}             
\usepackage{tikz}
\usetikzlibrary{decorations.markings, calc, shadings}
\usepackage{wrapfig}
\usepackage{menukeys}
\usepackage{bm}
\usetikzlibrary{positioning}
\usetikzlibrary{arrows.meta}
\usepackage{booktabs}
\usepackage{placeins}

\usepackage{amsthm}

\newtheorem{theorem}{Theorem}

\title{Hierarchical Behaviour Spaces}

\setrunningtitle{Hierarchical Behaviour Spaces}

\author{Michael Tryfan Matthews\textsuperscript{1,2}, Anssi Kanervisto\textsuperscript{1}, Jakob Foerster\textsuperscript{1,2}, Pierluca D'Oro\textsuperscript{1}, Scott Fujimoto\textsuperscript{1}, Mikael Henaff \textsuperscript{1}}

\emails{\{michael.tryfan.matthews,mikael314\}@gmail.com}

\affiliations{
$^{1}$\textbf{Meta Superintelligence Labs}\\
$^{2}$\textbf{University of Oxford}\\
}

\contribution{
We introduce Hierarchical Behaviour Spaces (HBS), a scalable hierarchical algorithm in which the controller commands options by specifying the coefficients of a linear combination over predefined reward functions.
    }
    {
    Scalable Option Learning~\citep{henaff2025scalable} was recently proposed as an approach to online hierarchical reinforcement learning that makes effective use of GPU parallelisation. We extend this approach from considering individual options specified by reward functions to the space of behaviours induced by the linear combination of these reward functions. Linear combinations of option reward functions have been investigated before, most notably in ~\citet{barreto2019option}.
    }

\contribution{
    We evaluate HBS on the NetHack Learning Environment~\citep{kuttler2020nethack}, showing strong performance and outperforming all baselines in the setup we consider. Although measuring progress on the benchmark is somewhat subjective, HBS shows what we believe to be the best known visitation rates of various notable waypoints in the early game, which may be an indicator of progress.
    }
    {
    Despite years of existence, NetHack has proven a remarkably resilient benchmark, with no approach making significant progress~\citep{hambro2022insights,piterbarg2023nethack,klissarov2023motif,klissarov2024maestromotif,paglieri2024balrog}. Win rate is not a meaningful metric, as to the best of our knowledge no artificial agent has ever beaten the game. There is ongoing debate on how best to measure progress on the benchmark, with the initial proposed metric of in-game score now widely seen as flawed~\citep{hambro2022dungeons,paglieri2024balrog,matthews2026revisiting}.
    }


\summary{Recent work in hierarchical reinforcement learning has shown success in scaling to billions of timesteps when learning over a set of predefined option reward functions. We show that, instead of using a single reward function per option, the reward functions can be effectively used to induce a space of behaviours, by letting the controller specify linear combinations over reward functions, allowing a more expressive set of policies to be represented. We call this method Hierarchical Behaviour Spaces (HBS). We evaluate HBS on the NetHack Learning Environment, demonstrating strong performance. We conduct a series of experiments and determine that, perhaps going against conventional wisdom, the benefits of hierarchy in our method come from increased exploration rather than long term reasoning.}

\begin{document}

\maketitle

\begin{abstract}

Recent work in hierarchical reinforcement learning has shown success in scaling to billions of timesteps when learning over a set of predefined option reward functions. We show that, instead of using a single reward function per option, the reward functions can be effectively used to induce a space of behaviours, by letting the controller specify linear combinations over reward functions, allowing a more expressive set of policies to be represented. We call this method Hierarchical Behaviour Spaces (HBS). We evaluate HBS on the NetHack Learning Environment, demonstrating strong performance. We conduct a series of experiments and determine that, perhaps going against conventional wisdom, the benefits of hierarchy in our method come from increased exploration rather than long term reasoning.

\end{abstract}

\section{Introduction}


Learning in long-horizon environments is a central problem in training reinforcement learning (RL) agents. Hierarchical RL, where decision making is done at multiple levels of abstraction, has been proposed as a solution to this issue \citep{sutton1999between, klissarov2025discovering}. Intuitively, we as humans do not consider our lives as a long sequences of individual muscle twitches, but employ abstraction and hierarchy when considering long term goals. For instance, when deciding on impactful life decisions like moving country or choosing a career, we would typically consider how it would affect high-level concepts like our quality of life over many years, rather than considering in turn how each day would be affected. Despite this seemingly intuitive mapping to human behaviour, hierarchical RL has shown limited practicality in online settings, where \textit{flat} (i.e. non-hierarchical) methods are still overwhelmingly the tool of choice.

Rather than trying to learn hierarchy end-to-end, recent work~\citep{henaff2025scalable} has proposed making use of a set of predefined reward functions to guide the learning of low-level policies, with the high level controller policy then learning how to sequence these. While this has shown success, the agent is limited in its expressivity to picking one of the predefined option rewards at any given time.

To help overcome this limitation, we propose allowing the controller to linearly interpolate between the reward functions. This increases the expressivity of low-level policies that can be induced, as even simply interpolating between 2 reward functions can induce arbitrary behaviour that is not present at either extremes. As the dimensionality of the behaviour space increases this effect only grows. We call this method Hierarchical Behaviour Spaces (HBS).


We evaluate our method in the challenging NetHack Learning Environment (NLE) \citep{kuttler2020nethack}: an unsolved benchmark based on the game of NetHack that requires reasoning over very long horizons to solve. We show that HBS outperforms prior work in the NLE and makes material progress on the benchmark, reaching various parts of the game with a consistency higher than any prior work, to the best of our knowledge.

\begin{figure}
\centering
\begin{tikzpicture}[scale=2.5]

  \tikzset{
    vertex/.style={circle, fill=black, inner sep=1.5pt},
    point/.style={circle, fill=blue!70!black, inner sep=1.2pt},
    edge/.style={thick, color=black!80},
    label font/.style={font=\small\itshape},
    arrow style/.style={
      -latex,
      thick,
      color=red!70!black,
      shorten <=2pt,
      shorten >=2pt,
    },
  }

  \shade[top color=blue!8, bottom color=blue!25, opacity=0.6]
    (0, 1.8) 
    .. controls (0.6, 1.9) and (1.4, 1.4) .. (1.7, 0.5)
    .. controls (1.3, -0.1) and (0.5, -0.3) .. (-0.6, 0.1)
    .. controls (-0.1, 0.7) and (0.3, 1.3) .. (0, 1.8);

  \draw[edge] (0, 1.8) 
    .. controls (0.6, 1.9) and (1.4, 1.4) .. (1.7, 0.5);
  \draw[edge] (1.7, 0.5) 
    .. controls (1.3, -0.1) and (0.5, -0.3) .. (-0.6, 0.1);
  \draw[edge] (-0.6, 0.1) 
    .. controls (-0.1, 0.7) and (0.3, 1.3) .. (0, 1.8);

  \node[vertex, label={[label font]above:{$\mathcal{R}_1$}}]  (A) at (0, 1.8)   {};
  \node[vertex, label={[label font]right:{$\mathcal{R}_2$}}]   (B) at (1.7, 0.5) {};
  \node[vertex, label={[label font]left:{$\mathcal{R}_3$}}]    (C) at (-0.6, 0.1) {};

  \node[point, label={[label font, text=blue!70!black]below:{$a \cdot \mathcal{R}_1 + b \cdot \mathcal{R}_2 + c \cdot \mathcal{R}_3$}}] 
    (p) at (0.55, 0.75) {};


  \node[font=\Large, anchor=east] (piOmega) at (-1.3, 0.75) {$\pi_{\Omega}(s)$};

  \node[font=\Large, anchor=west] (piomega) at (2.2, 0.75) 
    {$\pi_{\omega}(s,\rho)$};
  \coordinate (oTarget) at ([xshift=-5pt,yshift=3pt]piomega.east);

  \draw[arrow style] 
    (piOmega.east) to[out=10, in=170] (p.west);

  \draw[arrow style] 
    (p.east) to[out=-5, in=90] (oTarget);

\end{tikzpicture}
\caption{The $n$ reward functions induce a $n-1$ simplex of behaviours. The controller $\pi_\Omega$ selects a linear combination of reward functions from the simplex, which the intra-option policy $\pi_\omega$ then acts to maximise for some number of timesteps. In this way, a large diversity of behaviours can be extracted from only a few reward functions.}
\label{fig:simplex}
\end{figure}
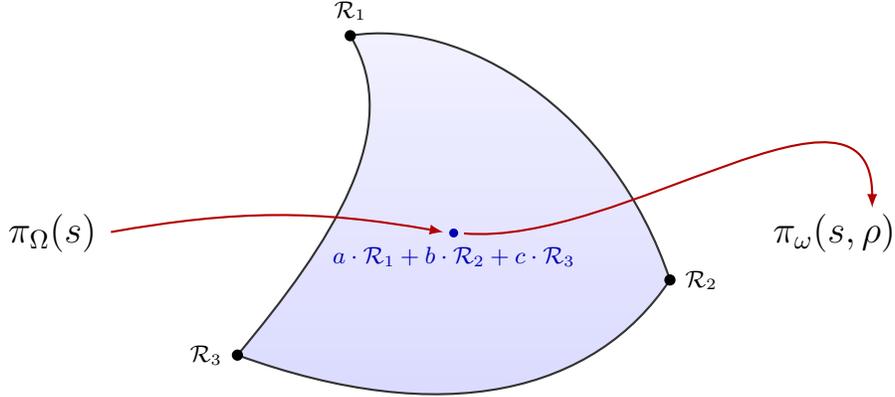

\section{Background}

\subsection{Reinforcement Learning}
We consider the standard RL formulation~\citep{sutton1998reinforcement}, with a Markov Decision Process (MDP) $M$ defined as $M = \langle \mathcal{S}, \mathcal{A}, \mathcal{R}, \mathcal{T}, \gamma\rangle$, where $\mathcal{S}$ is the set of states; $\mathcal{A}$ is the set of actions; $\mathcal{T}: \mathcal{S} \times \mathcal{A} \to \Delta (\mathcal{S})$ is the transition function, $\mathcal{R}: \mathcal{S} \to \mathbb{R}$ is the reward function and $\gamma$ is the discount factor. At each time step $t$, a state $s_t \in \mathcal{S}$ is given to the agent who chooses an action $a_t$ in response, causing the environment to transition to a new state $s_{t+1} \sim \mathcal{T}(\cdot | s_t, a_t)$ and a reward $R_t = \mathcal{R}(s_t, a_t)$ to be given to the agent. The goal of the agent is to learn a policy $\pi: \mathcal{S} \rightarrow \Delta(\mathcal{A})$ which maximises its expected discounted sum of rewards $\mathbb{E}_{\pi}[\sum_{t=0}^\infty \gamma^t R_t]$. 

\subsection{Options Framework}

The options framework~\citep{sutton1999between} extends the classic RL formulation with the notion of temporally extended behaviours or \textit{options}. A policy-over-options or \textit{controller} policy $\pi_\Omega: \mathcal{S} \to \Delta(\Omega)$ selects which option $\omega = (\pi_\omega, \beta_\omega)$ to run from the set of options $\Omega$, surrendering control to the intra-option policy $\pi_\omega: \mathcal{S} \times \Omega \to \Delta(\mathcal{A})$, which executes primitive actions until the termination function $\beta_\omega: \mathcal{S} \times \Omega \to \Delta(\{0, 1\})$ ends the option. After termination, control is passed back to the controller policy, which selects a new option to execute, and the process continues. Some works additionally define initiation sets $\mathcal{I}_\omega$ which restrict the set of states where an option can be initiated, but we consider the case where $\mathcal{I}_\omega = \mathcal{S}$, meaning all options can be initiated from any state.

A set of options $\Omega$ induces a semi-MDP (SMDP) on $M$ which we denote $M_\Omega$. A SMDP is analogous to an MDP where options replace actions and the transition function $\mathcal{T}_\Omega: \mathcal{S} \times \Omega \rightarrow \Delta(\mathcal{S})$ is given by the probability of reaching $s'$ upon termination when executing option $\omega$ starting from $s$. The reward for a given transition $(s, \omega, s')$ is the summed task reward over the course of $\pi_\omega$'s execution starting at $s$ and ending at $s'$.  

\subsection{Scalable Option Learning}

We focus on the setting proposed in~\citet{henaff2025scalable}, where both the controller and intra-option policy are learned concurrently, with each option associated with a reward function $\mathcal{R}^\omega: \mathcal{S} \times \Omega \to \mathbb{R}$. This reward can either be given a priori or learned using some form of reward synthesis, we focus on the former in our experiments. Each intra-option policy $\pi_\omega$ is trained to maximise the cumulative expected return under its reward function $\mathbb{E}_{R^\omega_t \sim \pi_\omega(\cdot, \omega)}[\sum_{t=0}^\infty \gamma^t R^\omega_t]$, under the assumption that it will be active for the rest of the episode. In this sense, each option is trained solely to optimise its own reward function in isolation, without regard to other options or the controller policy, in contrast to prior work~\citep{bacon2017option} which optimize both option and controller policies to maximise the task reward alone. The controller policy optimizes cumulative expected return over the task reward function $\mathcal{R}_\Omega: \mathcal{S} \to \mathbb{R}$. Rather than learning a termination function $\beta$, the controller decides the length for which the option should run for by acting in an augmented option space $|\mathcal{R}| \times |\mathcal{L}|$ where $\mathcal{L}$ is the set of valid option lengths which in practice are exponentially spaced ($\mathcal{L} = \{1, 2, 4, 8, 16..., 128\}$). In the rest of the manuscript we omit the option length action for clarity. 

\section{Hierarchical Behaviour Spaces}

\subsection{Motivation} \label{sec:motivation}

We first discuss some theoretical results from \citet{pmlr-v54-fruit17a} that motivate our algorithm. 

\begin{theorem}
    Let $M$ be an MDP, $\Omega$ a set of options, and $M_\Omega$ the corresponding SMDP. Let $\pi_\Omega$ be any stationary policy on $M_\Omega$ and $\mu$ the induced low-level policy on $M$. 
Let $A$ be any learning algorithm operating in the SMDP $M_\Omega$, let $m$ be the number of option calls executed in $M_\Omega$ with execution lengths $l_1, l_2, \dots, l_m$, and let $T_m = \sum_{i=1}^m l_i$ be the corresponding number of steps executed in the original MDP $M$. The following relationship then holds:
\begin{equation}
    \mathrm{Regret}(M, A, T_m) = \mathrm{Regret}(M_\Omega, A, m) + T_m (V^\star_M - V^\star_{M_\Omega})
\end{equation} \label{eq:regret}
where $V^\star_M$ and $V^\star_{M_\Omega}$ denote the maximum value in the original MDP and SMDP respectively.
\end{theorem}

This result says that when learning a controller in the high-level SMDP, the regret in the original MDP $M$ over a time horizon $T_m$ can be decomposed into the sum of two terms: the regret in the SMDP $M_\Omega$, and the gap between the optimal value in the original MDP and the optimal value in the SMDP. The first term reflects the learning speed in the temporally compressed SMDP, which will typically be faster than in the original MDP due to its shorter horizon $(m \ll T_m)$. The second term arises when the set of option policies is not expressive enough to fully represent the optimal policy $\pi^\star$ in the original MDP. That is, if at some state $s$ visited by the agent, there are no option policies $\pi_\omega$ which match $\pi^\star$ for the next few steps, performance may suffer in the original MDP despite the controller being optimal in the SMDP.

This result highlights the need for a set of option policies that is sufficiently expressive to represent the optimal policy as closely as possible. Given a set of $n$ reward functions, where $n$ may be small, SOL~\citep{henaff2025scalable} will extract $n$ options, which may be insufficiently expressive. We consider instead using these reward functions to induce a \textit{space of behaviours} by considering linear combinations of them, which may be far more expressive.



\subsection{Method}

We propose Hierarchical Behaviour Spaces (HBS), a two-level hierarchical RL algorithm in which the controller commands behaviours induced from a simplex of predefined reward functions $\{\mathcal{R}_1, ..., \mathcal{R}_n\}$.

HBS jointly trains a controller $\pi_\Omega: \mathcal{S} \to \Delta([0,1]^n)$ that acts in a temporally compressed process by specifying a vector of reward coefficients $\rho \in [0,1]^n$, and a behaviour-conditioned policy $\pi_\omega: \mathcal{S} \times [0,1]^n \to \Delta(\mathcal{A})$ that outputs primitive actions conditioned on the reward coefficients output by the controller  (Figure \ref{fig:simplex}). 

The two policies act on different time scales and are trained to optimise their respective cumulative discounted return with different discount factors:
%

\begin{align}
    \pi_\Omega^* &= \text{argmax}_{\pi_\Omega} \mathbb{E}_{R_t \sim \pi_\Omega}\left[\sum_{t=0}^\infty \gamma_\Omega^t R_t\right] \\
    \pi_\omega^*(\cdot, \rho) &= \text{argmax}_{\pi_\omega} \mathbb{E}_{R_t \sim \pi_\omega(\cdot, \rho)}\left[\sum_{t=0}^\infty \gamma_\omega^t \cdot \rho^T (\mathcal{R}^1_t, \dots,\mathcal{R}^n_t) \right]
\end{align}
Where $\rho^T (\mathcal{R}^1_t, \dots,\mathcal{R}^n_t)$ is the linear combination of the reward functions weighted by the behaviour vector $\rho$.

HBS can be seen as a generalization of SOL to allow for interpolation between the various reward functions, rather than simply having to pick one. This greatly expands the range of possible behaviours that the controller can induce, allowing us to reduce the second term in Equation \ref{eq:regret}.

From a practical perspective, this entirely changes how we should think about the set of reward functions. In SOL each reward function $\mathcal{R}_i$ can be paired with a policy $\pi_i$ that maximises it, with these then being sequenced by the controller. Each $\pi_i$ should be a self-contained sub-policy that is optimal for given sub-trajectories. In contrast, the reward functions in HBS define \textit{axes of behaviour}. There is no requirement for any one reward function to induce a sensible policy by itself, rather each reward function should ideally increase the set of representable behaviours the controller can induce, by being orthogonal to the other reward functions in behaviour space.

As with SOL, both the controller and intra-option policies are implemented as separate heads on top of a shared network. We kept the other design choices which enable high throughput, such as parallelised advantage and return computations, allowing us to scale this method to run for billions of timesteps. 

In practice, we found that using discrete, quantised bins for the controller to specify the coefficients outperformed using a continuous action space, even though it reduces the expressivity of policies that can be induced, in line with prior work~\citep{farebrother2403stop}. We also found that \textit{not} normalising the reward coefficients by the sum of the magnitudes marginally improved performance, even though this means that redundant behaviour vectors can be specified.



\section{Experiments}

\subsection{Experimental Setup}

We evaluate our method on the NetHack Learning Environment (NLE, \citet{kuttler2020nethack}): a challenging, unsolved benchmark that requires reasoning over very long time horizons\footnote{The average successful human ascensions take $10^4-10^5$ turns~\citep{hambro2022dungeons, paglieri2024balrog}.}. We implement HBS using the SOL~\citep{henaff2025scalable} codebase, which is itself based off the PPO~\citep{schulman2017proximal} implementation from Sample Factory~\citep{petrenko2020sample}.

We follow \citet{matthews2026revisiting} in our setup, meaning that unlike most prior work on the NLE, we make use of the full action space. This actually somewhat alleviates the need for long term credit assignment, as the full action space allows the agent to induce a limited set of useful macro-actions that can span over many in-game steps. For instance, the agent can (and indeed does learn to) execute the actions \keys{9}, \keys{9}, \keys{s}, which causes the agent to wait and heal for 99 timesteps. Conversely, the agent can also take actions that do not step the in-game turn counter, for instance by navigating around menus. The MDP that the agent acts in does not therefore perfectly align with the underlying game and we make a distinction between \textit{timesteps} (measure of time in MDP) and \textit{turns} (measure of time in NetHack).

In further contrast to prior work on the NLE, we use the \textit{scout} rather than \textit{score} as the reward function we seek to maximise, as recommended by \citet{matthews2026revisiting}. As has been noted many times in prior work~\citep{kuttler2020nethack,hambro2022insights,hambro2022dungeons,wolczyk2024fine,paglieri2024balrog,matthews2026revisiting}, the in-game score is a flawed metric for assessing progress on the NLE, as it is easily gameable by staying on the first dungeon level and killing weak enemies as they spawn, without making any actual progress through the game. Scout reward is increased by revealing new parts of the dungeon map. Since the in-game map is finite, maximising scout reward will lead the agent to the final level, whereas score is unbounded and can be infinitely maximised while never leaving the first level.

\begin{figure}
    \centering
    \includegraphics[width=0.9\linewidth]{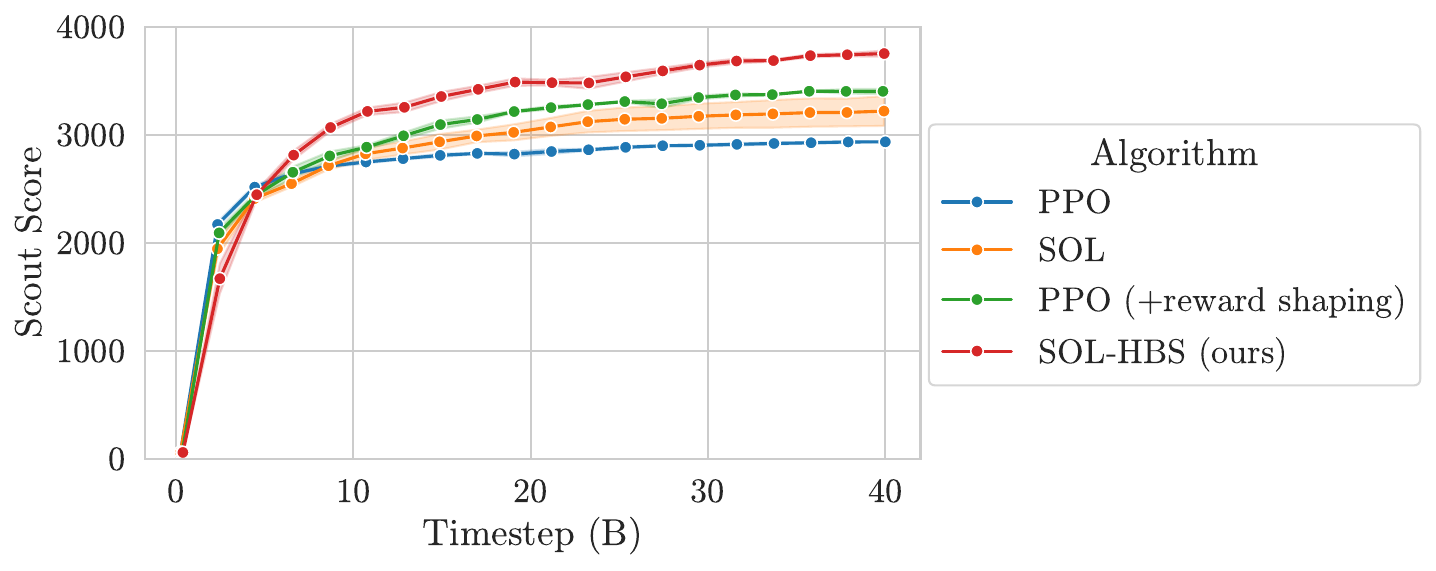}
    \caption{Results on the NetHack Learning Environment. The shaded areas denote 1 standard error over 5 seeds. We show the best setting for each method: for HBS this is using the full space of 5 reward functions, while for SOL and PPO with reward shaping this is just using the scout and $-\bm{\Delta}(\textbf{dlvl})$ rewards.}
    \label{fig:main_results}
\end{figure}

Along with the scout reward, we choose 4 other reward functions to define the behaviour space:
\begin{itemize}
    \item $-\bm{\Delta}(\textbf{dlvl})$: A reward for decreasing the agent's dungeon level. While progress is generally made by increasing the dungeon level, it is often useful to return to prior levels, for example to retreat or navigate between dungeon branches. 
    \item $-\bm{\Delta}(\textbf{AC})$: A reward for decreasing armour class, corresponding to better armour.
    \item $+\bm{\Delta}(\textbf{Food})$ A reward for eating food. Hunger increases every turn and starvation is a common mode of death.
    \item $+\bm{\Delta}(\textbf{XP})$ A reward for increasing experience, which is mostly gained by killing enemies.
\end{itemize}

These rewards form a 5-dimensional behaviour space for HBS to operate in. We also consider using these rewards directly as options for SOL, as well as simply adding them to PPO as intrinsic rewards. As with the evaluation in \citet{henaff2025scalable}, we are unable to practically compare to other hierarchical methods, as we cannot feasibly run them for the billions of timesteps required by the NLE. We detail the hyperparameters used in Appendix \ref{app:hyps}.

\subsection{Results}

The results in the NLE are shown in Figure \ref{fig:main_results} where we run for 40 billion timesteps. We see that, while HBS is less sample efficient at first, it begins outperforming the baselines at around 5 billion timesteps and converges to a higher return. Notably, this is in contrast to prior work in HRL which has posited sample efficiency as a benefit of hierarchy~\citep{klissarov2025discovering}.

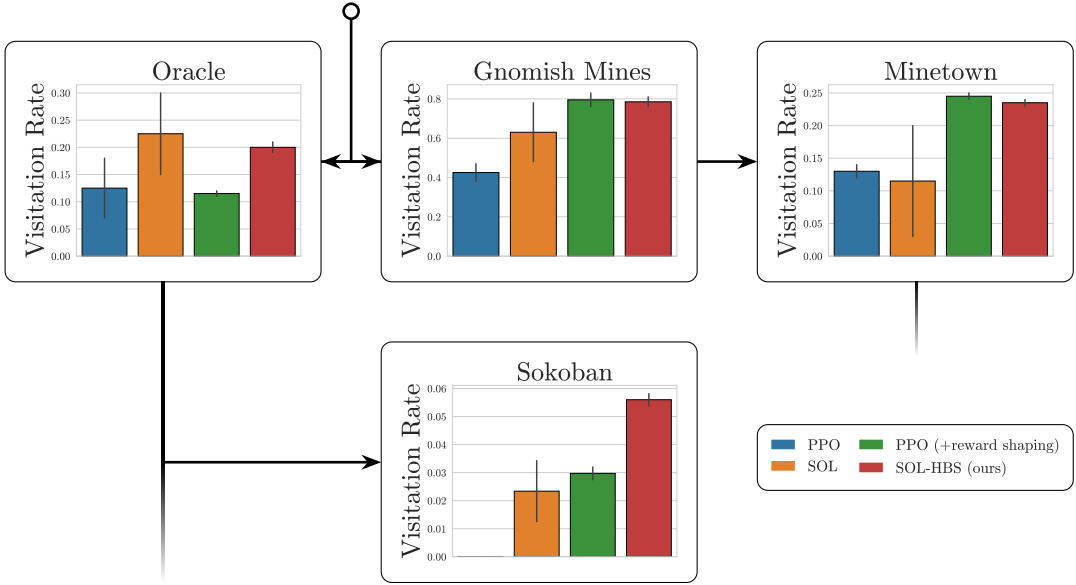
\begin{figure}
\centering
\begin{tikzpicture}[
  node distance=14mm and 22mm,
  every node/.style={inner sep=0pt, outer sep=0pt},
  plot/.style={
    fill=white!10,
    draw=black!40!black,
    rounded corners,
    inner sep=1mm
  },
  >=Stealth,
  every path/.style={draw, -{Stealth}} 
]

\node[plot] (C) {\includegraphics[width=4cm]{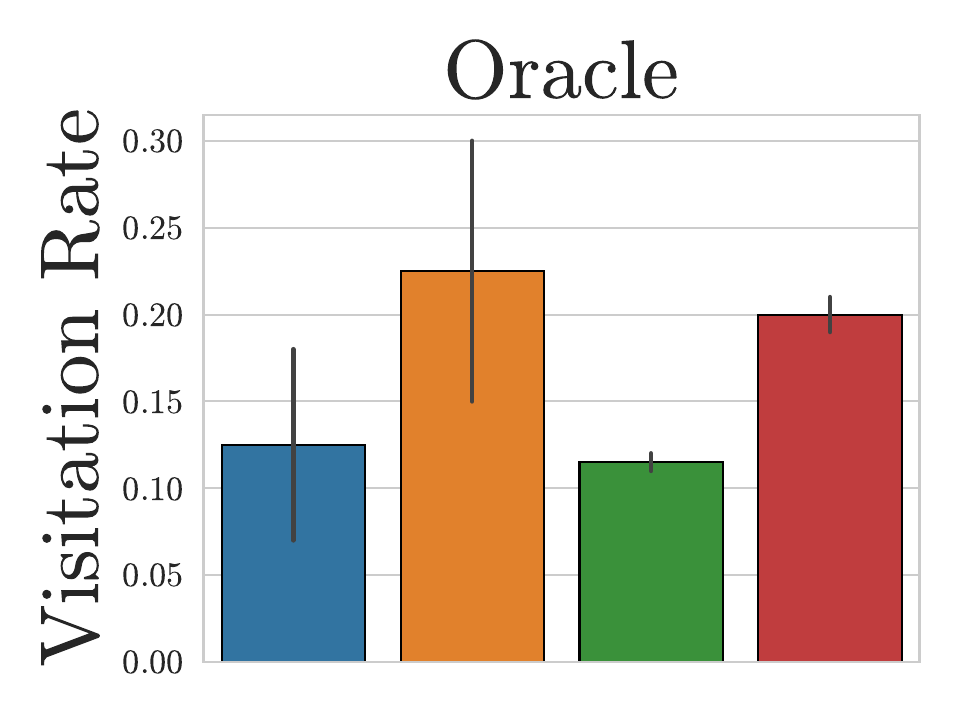}};
\node[plot] (A) [right=8mm of C] {\includegraphics[width=4cm]{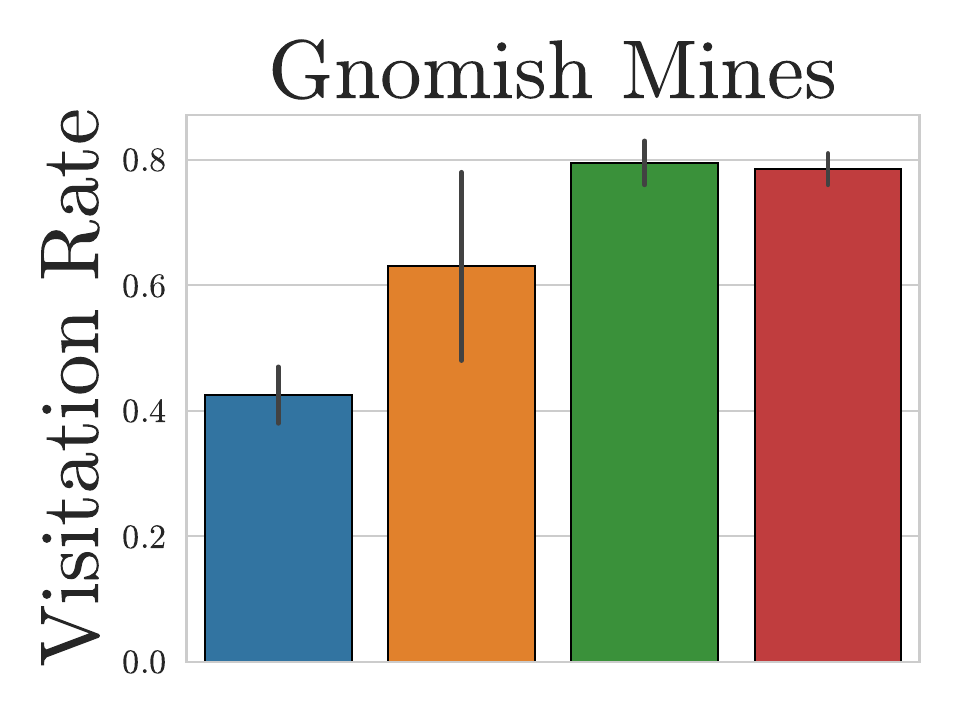}};
\node[plot] (B) [right=8mm of A] {\includegraphics[width=4cm]{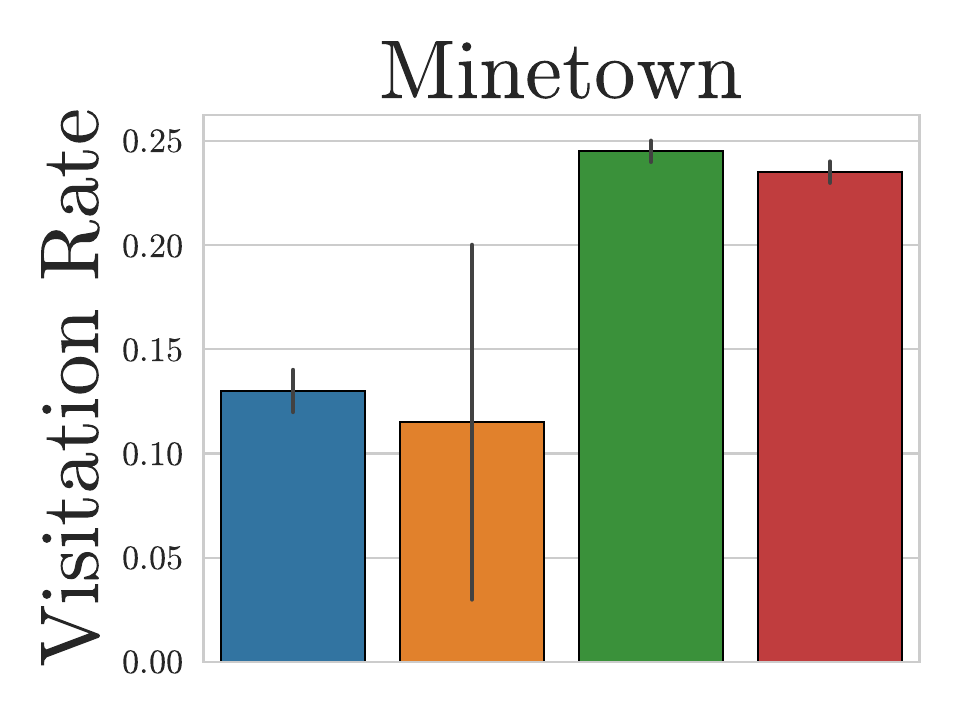}};
\node[plot] (D) [below=8mm of A] {\includegraphics[width=4cm]{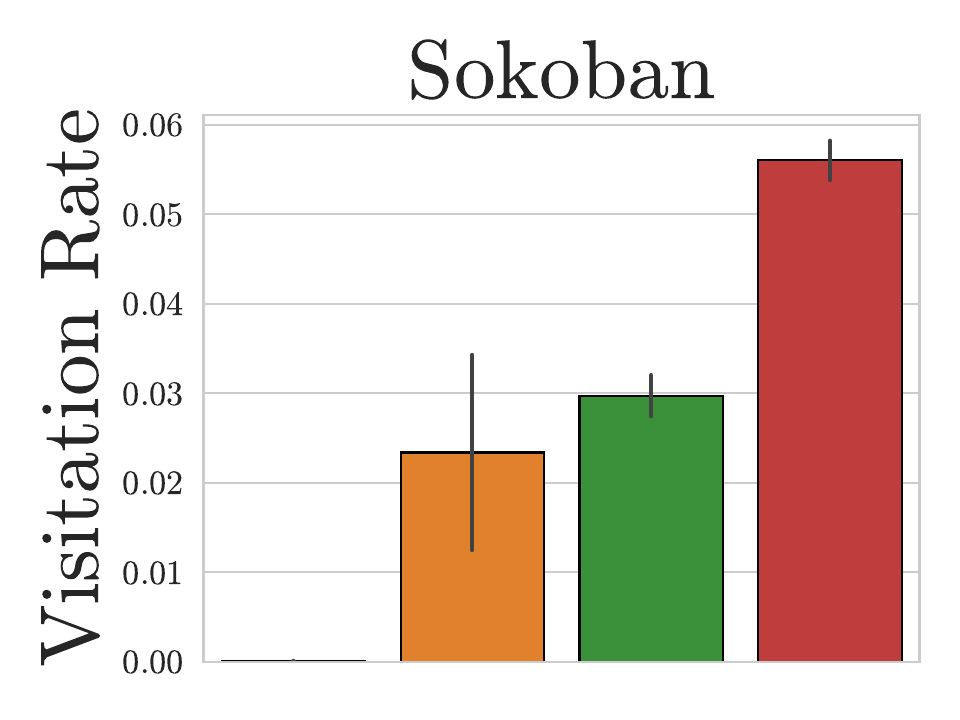}};


\node[plot, draw=black]
  [below=19mm of B] (L)
  {\includegraphics[width=4cm]{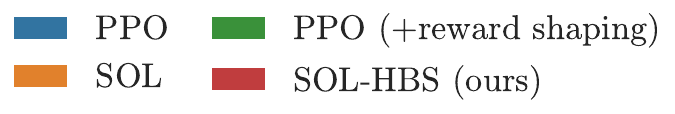}};


\draw[line width=1pt, -] ($0.5*(C.east)+0.5*(A.west)$) -- ++(0,20mm)
    node[circle, draw=black, fill=white, inner sep=2pt] {};

\draw[line width=1pt] (C.east) -- (A.west);
\draw[line width=1pt] (A.west) -- (C.east);
\draw[line width=1pt] (A.east) -- (B.west);

\draw[line width=1pt] (C.south) |- (D.west);

\shade[shading=axis, top color=black, bottom color=white, draw=none]
  ([xshift=-0.5pt, yshift=-24mm]C.south) rectangle ++(1pt,-16mm);

\shade[shading=axis, top color=black, bottom color=black, draw=none]
  ([xshift=-0.5pt, yshift=-0mm]C.south) rectangle ++(1pt,-24mm);

\shade[shading=axis, top color=black, bottom color=white, draw=none]
(B.south) rectangle ++(1pt,-10mm);

\end{tikzpicture}
\caption{Visitation rates of early waypoints in NetHack for each method. The arrows indicate the underlying structure of the waypoint locations in the game. The error bars show 1 standard error over 5 seeds. HBS achieves high visitation rates on both branches of the dungeon, unlike PPO which tends to favour the Gnomish Mines and SOL which, depending on the seed, will favour one of the branches but not generally learn to navigate between them.}

\label{fig:waypoints}
\end{figure}

Taking a closer look at individual milestones in the NLE (Figure \ref{fig:waypoints}), we can see the qualitative differences between the methods. The Gnomish Mines form the first branch off from the main dungeon in NetHack, with the entrance somewhere on dungeon levels 2--4. We see that vanilla PPO enters them on less than half of all episodes. This is likely due it implementing a strategy to always descend and never revisit previous dungeon levels or different branches. This means that if the agent descends past the entrance to the mines by continuing down the main dungeon, then it will never go back to find it. In contrast, we see that SOL and especially HBS and PPO with the $-\bm{\Delta}(\textbf{dlvl})$ intrinsic reward enter the mines far more regularly, indicating that those agents have learned to traverse back to previous dungeon levels.

The Oracle appears between levels 5 and 9 in the main branch. We see both variants of PPO visit it quite rarely, while SOL visits it frequently on some seeds but infrequently on others. HBS is the only agent to show strong visitation for both branches of the dungeon.

We also see that all agents except vanilla PPO start making progress on finding the Sokoban branch of the dungeon, which appears one level beneath the Oracle, with HBS finding it roughly twice as often as the next best agent. As well as simply requiring the agent to survive for many more floors, this branch is entered on an up-stair and therefore requires the agent to have discovered that going back up the dungeon can sometimes be beneficial.

\section{Discussion}


\subsection{Does HBS allow reasoning over long horizons?}

Long term reasoning is often invoked as a motivation behind HRL methods, but it is not clear for our case, whether this is backed up by evidence. 

First off we note that increasing the discount factor $\gamma$ as high as $0.9995$ for a simple PPO agent on the NLE produces monotonically increasing returns, before performance collapses at higher $\gamma$, likely due to the increased variance of return estimation. This indicates that the ability to act with respect to long term rewards is important in the NLE, as one would intuitively expect.

\begin{figure}
    \centering
    \includegraphics[width=0.85\linewidth]{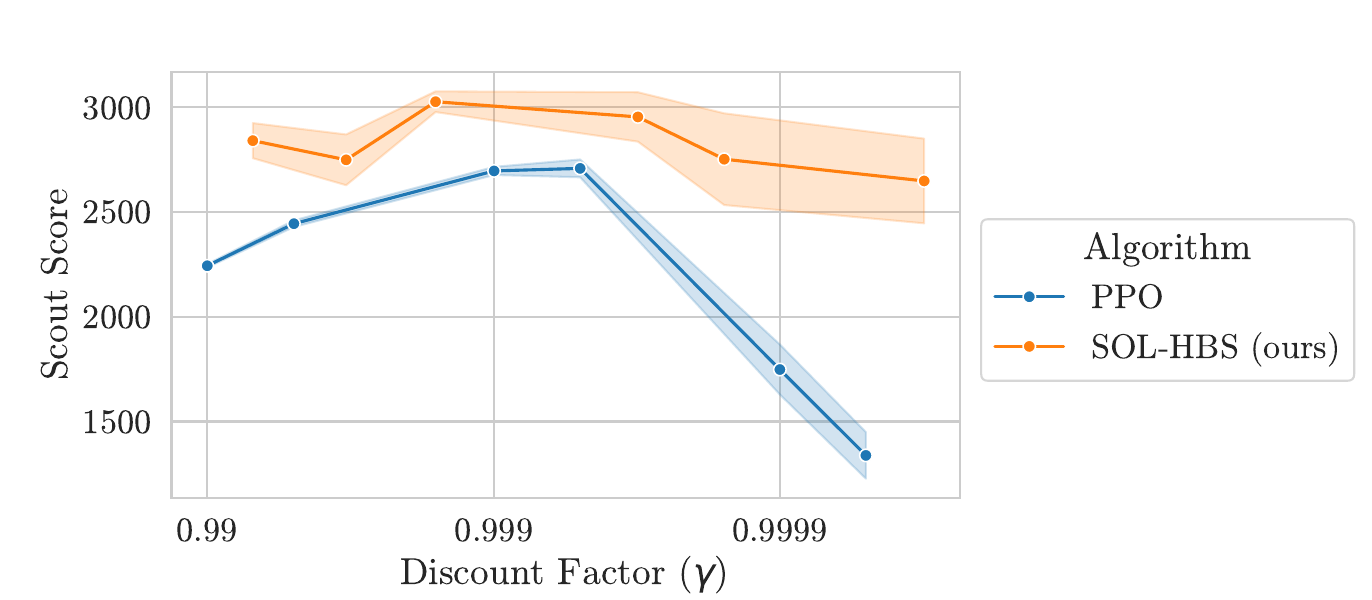}
    \caption{Results on the NetHack Learning Environment for PPO and HBS trained for 8 billion timesteps with different discount factors. The shaded area denotes 1 standard error over 5 seeds. For HBS we vary the controller discount factor $\gamma_\Omega$, while keeping the intra-option discount factor $\gamma_\omega$ fixed at $0.999$. PPO performs well at intermediate values but performance collapses with high discount factors. HBS is largely unaffected by the discount factor of the controller.}
    \label{fig:gamma}
\end{figure}

We then consider HBS, where we fix the intra-option discount factor $\gamma_\omega=0.999$ but vary the controller discount factor $\gamma_\Omega$. Note that $\gamma_\Omega$ corresponds to discounting in the primitive MDP, so it can be directly compared to $\gamma_\omega$ and $\gamma$.  We see that $\gamma_\Omega$ has a limited effect on the performance, and perhaps surprisingly, peaks at a similar point to $\gamma$ for PPO (Figure \ref{fig:gamma}).

This result challenges the intuition that the benefits of HBS come from the ability for the controller to reason over longer timescales, as we see that the best performance comes roughly when $\gamma_\Omega = \gamma_\omega$ (note we are not claiming this is a general rule, it is just what we observe in this limited experiment). Interestingly, HBS with a low $\gamma_\Omega$ close to $0.99$ still outperforms PPO with a $\gamma$ of $0.9995$.

This raises the obvious question: if HBS does not improve long term credit assignment, then why does it outperform other methods?

\subsection{HBS as automated tuning of intrinsic rewards}

An alternative way to view HBS is as a method for automatic tuning of intrinsic rewards. Consider simply using our predefined set of rewards as intrinsic bonuses for a flat agent, as we did for a baseline. Performing a hyperparameter sweep over every combination of coefficients scales exponentially with the number of reward functions and quickly becomes intractable. HBS can be seen as a method for automatically finding the best coefficients for each intrinsic reward.

But HBS doesn't simply set the coefficients once, it rather dynamically modifies them throughout the episode. This could be seen as expanding the expressivity of intrinsic rewards from static to dynamic bonuses, in effect massively increasing the set of intrinsic reward functions we are optimising over.

\begin{figure}
    \centering
    \includegraphics[width=0.9\linewidth]{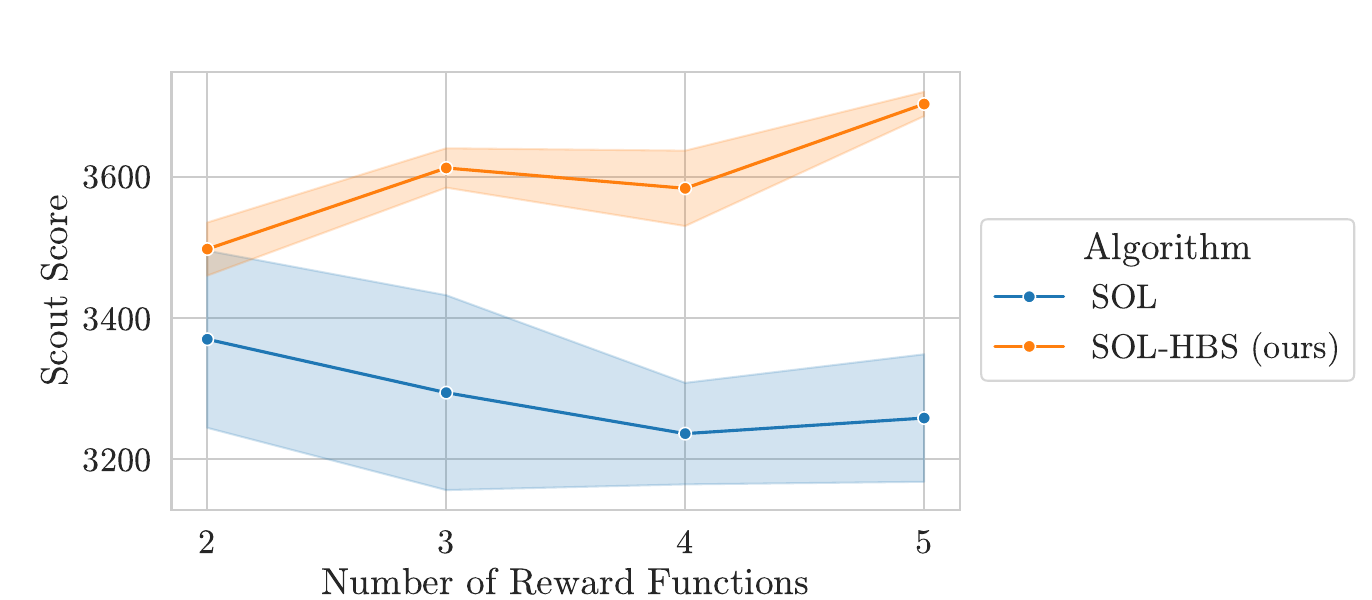}
    \caption{Performance on the NLE as the option space grows for both SOL and HBS at 40 billion timesteps. The shaded area denotes 1 standard error over 5 seeds. The agents start with just the scout and $-\bm{\Delta}(\textbf{dlvl})$ rewards, before $+\bm{\Delta}(\textbf{Food})$, $-\bm{\Delta}(\textbf{AC})$ and $+\bm{\Delta}(\textbf{XP})$ are added in that order. We see that HBS can make use of the extra axes of behaviour given by the new reward functions, in contrast to SOL whose performance degrades with more options.}
    \label{fig:options}
\end{figure}

This alleviates much of the burden from the RL practitioner, who can now simply specify a set of intrinsic rewards that might be useful (and maybe even only some of the time) and let HBS figure out which ones to apply and when. A key indicator of whether this claim is manifested in reality is how performance is impacted as we add more intrinsic rewards to the mix. Figure \ref{fig:options} indeed shows that as we add to the set of reward functions, HBS performance does generally increase, in contrast to SOL where performance decreases.

This implies that the performance improvements from the increased expressivity of HBS could come from enhanced exploration rather than any benefits to long term credit assignment. 

\section{Related Work}

\subsection{Hierarchical RL}

While early work in Hierarchical RL~\citep{dayan1992feudal,kaelbling1993learning,sutton1999between,precup2000temporal} often focused on predefined options~\citep{kaelbling1993learning,sutton1999between,precup2000temporal}, the trend in later work has been to try and learn options end-to-end in service of a single task reward either through subgoals~\citep{mcgovern2001automatic,stolle2002learning,menache2002q}, arbitrary learned options~\citep{bacon2017option} or options that themselves directly optimize the task reward~\citep{klissarov2017learnings, li2019sub, klissarov2021flexible}. Similar work has looked at extracting a diverse set of options from gathered data~\citep{gregor2016variational,eysenbach2018diversity,sharma2019dynamics} and from offline trajectories~\citep{pertsch2021accelerating,shi2022skill,park2023hiql,park2025horizon}. We consider the question of reward synthesis for options to be an important but orthogonal line of investigation to our work, where we focus solely on learning the best possible agent given a set of reward functions.

Most similar to our work is the Option Keyboard~\citep{barreto2019option}, which learns a hierarchical policy over linear combinations of `cumulants'. The Option Keyboard employs a two stage learning process, where value functions are first learned independently and linear combinations can then be synthesised zero-shot. We differ in adopting a jointly trained, coefficient-conditioned architecture that scales to environments like the NLE.

\subsection{NLE}

While much of the initial work on the NLE was done with tabula rasa RL~\citep{kuttler2020nethack,hambro2022insights}, a wide range of methods have been applied since. These include using offline data~\citep{hambro2022dungeons,piterbarg2023nethack,wolczyk2024fine}, LLM generated rewards~\citep{klissarov2023motif,klissarov2024maestromotif,zheng2024online}, LLM actors~\citep{paglieri2024balrog}, symbolic agents~\citep{hambro2022insights}, exploration bonuses~\citep{henaff2022exploration} and hierarchical agents~\citep{matthews2022hierarchical,klissarov2024maestromotif,henaff2025scalable}. As of this date and to the best of our knowledge, humans remain the only agents to have ever beaten the game.

\section{Conclusion}

In conclusion, we present HBS, a hierarchical RL algorithm in which a controller commands a linear combination of reward functions for an intra-option policy to follow. We show that this setup allows us to train a hierarchical agent with a greatly increased expressivity compared to prior work. We show state of the art results on the NLE, a challenging, unsolved benchmark that requires significant exploration. We hope that HBS can serve as a useful tool for RL practitioners working in long-horizon and hard exploration environments.







\bibliography{main}
\bibliographystyle{rlj}

\clearpage
\beginSupplementaryMaterials


\appendix

\section{Hyperparameters} \label{app:hyps}

For the base PPO implementation we use the same hyperparameters as SOL, except we increase $\gamma$ to $0.999$, and modify the number of works and environments, since we use a multi-GPU setup so can accommodate more parallel workers (Table \ref{tab:ppo_hyps}).

We also list the additional hyperparameters for SOL (Table \ref{tab:sol_hyps}) and SOL-HBS (Table \ref{tab:hbs_hyps}).

We list the hyperparameters for the intrinsic rewards we tried with PPO in Table \ref{tab:ppo_int_hyps}. We tried each intrinsic reward independently, as well as all together at once. Trying every combination would have been $3^4=81$ trials (each with 5 seeds, as the NLE has a high variance) and was impractical with our compute budget, especially considering these runs tended to take more than 10 billion timesteps each to converge.

\FloatBarrier
\begin{table}
    \begin{center}
    \scalebox{1.0}{
        \begin{tabular}{lll}
        \toprule
        \textbf{Hyperparameter} & \textbf{Value} \\
        \midrule
        Recurrence  & $256$ \\
        Normalise Returns  & False \\
        V-Trace  & True \\
        Num Workers  & $36$ \\
        Num Envs Per Worker  & $32$ \\
        Batch Size  & $32768$ \\
        Reward Scale  & $0.01$ \\
        Inventory Encoder  & Attention \\
        Inventory Query Heads  & $4$ \\
        Discount Factor ($\gamma$)  & $0.999$ \\
        Value Loss Coefficient & $0.5$ \\
        Exploration Loss Coefficient  & $0.003$ \\
        Reward Clip  & $10000$ \\
        Max Token Length  & $12$ \\
        Model  & SymbolicGlyphTokenNetEmbeddingBag \\
        Map Input Type  & Glyphs \\
        Inventory Input Type  & Tokens \\
        Crop Dimension  & $12$ \\
        Epochs  & $1$ \\
        Learning Rate  & $0.0002$ \\
        Num GPUs  & $4$ \\
        \bottomrule 
        \end{tabular}}
    \end{center}
    \caption{PPO hyperparameters. These were used for PPO, SOL and SOL-HBS.}
    \label{tab:ppo_hyps}
\end{table}

\begin{table}
    \begin{center}
    \scalebox{1.0}{
        \begin{tabular}{lll}
        \toprule
        \textbf{Hyperparameter} & \textbf{Value} \\
        \midrule
        Controller Exploration Scale  & $1$ \\
        Controller Reward Scale & $0.001$ \\
        Num Option Steps & Adaptive: $\{1, 2, 4, 8, 16, 32, 64, 128\}$ \\
        Reward Scale $-\bm{\Delta}(\textbf{dlvl})$ & 100 \\
        Reward Scale $-\bm{\Delta}(\textbf{AC})$ & 250 \\
        Reward Scale $+\bm{\Delta}(\textbf{Food})$ & 0.1 \\
        Reward Scale $+\bm{\Delta}(\textbf{XP})$ & 4 \\
        \bottomrule 
        \end{tabular}}
    \end{center}
    \caption{SOL additional hyperparameters. These were used for both SOL and SOL-HBS.}
    \label{tab:sol_hyps}
\end{table}

\begin{table}
    \begin{center}
    \scalebox{1.0}{
        \begin{tabular}{lll}
        \toprule
        \textbf{Hyperparameter} & \textbf{Value} \\
        \midrule
        HBS Coefficient Spacing  & Linear \\
        HBS Num Coefficients & $3$ \\
        HBS Normalise Coefficients & False \\
        \bottomrule 
        \end{tabular}}
    \end{center}
    \caption{SOL-HBS additional hyperparameters.}
    \label{tab:hbs_hyps}
\end{table}

\begin{table}
    \begin{center}
    \scalebox{1.0}{
        \begin{tabular}{lll}
        \toprule
        \textbf{Hyperparameter} & \textbf{Considered Values} & \textbf{Value} \\
        \midrule
        Reward Scale $-\bm{\Delta}(\textbf{dlvl})$ & $\{0, 50, 100\}$ & 100 \\
        Reward Scale $-\bm{\Delta}(\textbf{AC})$ & $\{0, 125, 250\}$ & 0 \\
        Reward Scale $+\bm{\Delta}(\textbf{Food})$ & $\{0, 0.05, 0.1\}$ & 0.1 \\
        Reward Scale $+\bm{\Delta}(\textbf{XP})$ & $\{0, 2, 4\}$ & 0 \\
        \bottomrule 
        \end{tabular}}
    \end{center}
    \caption{PPO Intrinsic Rewards.}
    \label{tab:ppo_int_hyps}
\end{table}



\end{document}